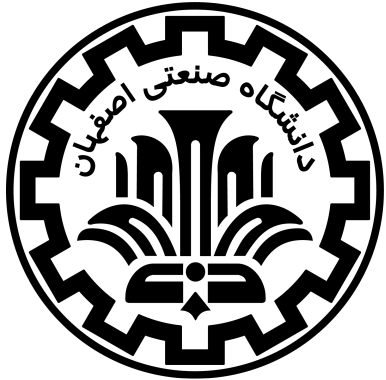

**ISFAHAN UNIVERSITY OF TECHNOLOGY**

**DEPARTMENT OF ELECTRICAL AND COMPUTER ENGINEERING**

# The Impact of Twitter Sentiments on Stock Market Trends

*Authors:*

Melvin Mokhtari (melvin.mokhtari@ec.iut.ac.ir)

Ali Seraj

Niloufar Saeedi

Adel Karshenas

## Authors' Declaration

We declare that the work included in the above paper is original and is an outcome of the research carried out by the authors indicated in it, and that we are the sole authors of this work. We would like to certify that there are no known conflicts of interest linked with this publication and that no financial assistance has been provided for this study that might have impacted its conclusion.

We also confirm that the work does not infringe on any copyrights or property rights of others, including licenses, and it is free from plagiarism. For the inclusion of figures, tables, animations, or text quotations from copyrighted works (including websites), necessary permission has been obtained from the copyright holders. By doing so, we guarantee that we have followed the regulations concerning intellectual property.

**Abstract**

The Web is a vast virtual space where people can share their opinions, impacting all aspects of life and having implications for marketing and communication. The most up-to-date and comprehensive information can be found on social media because of how widespread and straightforward it is to post a message. Proportionately, they are regarded as a valuable resource for making precise market predictions. In particular, Twitter has developed into a potent tool for understanding user sentiment.

This dissertation examines how well tweets can influence stock symbol trends. We analyze the volume, sentiment, and mentions of the top five stock symbols in the S&P 500 index on Twitter over three months. Long Short-Term Memory, Bernoulli Naïve Bayes, and Random Forest were the three algorithms implemented in this process. Our study revealed a significant correlation between stock prices and Twitter sentiment.

# Table of Contents

# List of Figures

# List of Tables

# Chapter 1

# Introduction

This chapter introduces the research's motivation, describes the problem, and defines the scope and theoretical approaches. The goals of the research are then presented. Finally, it highlights this dissertation's main contributions and outlines the document's structure.

## 1.1 Overview

The recent technological revolution has resulted in an unprecedented data deluge that has fundamentally altered how we view social and economic sciences. Computers and the Internet are now a part of almost everyone's daily lives. Similar increases in online activity have been sparked by the continuously growing Internet usage, such as for business or news. Due to this issue, social media's emergence has provided a platform for web users to express their opinions on various topics. Therefore, events and the creation of textual documents have grown significantly over the past few years on these platforms, and consequently, massive datasets are being produced by the interaction of technological systems, documenting collective behavior in previously unthinkable ways. Finally, we can discover the global population's interests, worries, and intentions regarding various economic, political, and cultural phenomena in this massive database of Internet activities.

Twitter is a widely used microblogging and social networking platform where people can read, write, publish, and update short text messages called tweets [1–4]. In 2021, Twitter's global user base was approximately 429.79 million. Furthermore, it is anticipated that by 2025, there will be 497.48 million users worldwide [5]. As a result, this microblogging and social networking service piqued the interest of researchers from a variety of fields, including politics, health care, and finance.

Sentiment analysis is the name given to the scientific study of the semantic content of these tweets. In general, sentiment analysis is a technique for identifying and classifying a text's polarity, aiming to determine whether a specific document has a positive or negative value in accordance with a standard categorization. According to some researchers, processing Twitter user opinions in the context of financial markets makes it possible for everyone to gather pertinent information about the stock market and use it to forecast changes in stock prices. Indeed, since the very beginning of stock markets, investors have always sought to gain a competitive advantage. Not illegally, such as through insider trading, but by wisely and sparingly using information.

Here, we present a case study on financial markets, one of the numerous fields in which data collection, analysis, and modeling are applied. According to our analysis, social network metrics on social aspects are particularly helpful in understanding financial turnovers. Indeed, collective phenomena such as herd behavior frequently precipitate financial crises and demonstrate the financial system's complex nature. As a result, policymakers are eager to anticipate unusual collective behavior among investors to be able to intervene sooner when necessary.

The inherent risks and complications of investing in the financial market have led to the development of techniques that minimize investor losses and maximize returns. These techniques use machine learning algorithms to find connections between individuals' thoughts about the financial market and the movements of stock prices. This will make it possible to predict how the market will change.

## 1.2   Motivation and Objectives

The impact of Twitter on financial markets is the focus of this dissertation. The research will particularly examine how tweets about a company may influence its share price trends.

Early attempts to predict the stock market aimed to determine if stock prices could be predicted. There are two approaches to this: the random walk theory and the efficient market hypothesis (EMH) [6]. According to the efficient market hypothesis, rather than reflecting current and historical prices, stock indices primarily reflect news from existing investors. However, the random walk theory contends that because news breaks at unpredictable times, predictions can never be trusted to be accurate.

According to research by Qian et al., EMH and the random walk model's basic assumptions have been identified and explained [7]. These theories have demonstrated that some degree of forecasting is feasible using different economic and business indicators. According to the widespread use of the powerful version of EMH, prices aggregate publicly available information and immediately reflect the most recent version [8]. It is widely acknowledged that news influences the macroeconomic movement of the markets. However, research indicates that social media buzz significantly impacts the microeconomic level, especially in the major indices like the S&P 500 and DJIA [9–12]. It has been found through earlier research that investor sentiments and decisions ultimately drive the market [7]. Therefore, a model that considers feelings will provide better microeconomic predictions.

Traders take positions in the market that are held for only a short time, sometimes for just a few moments, so many investors may quickly join and leave those positions daily. Consequently, many investors will look for anything that will give them an advantage when placing market bets. Several great traders have prepared bots based on artificial intelligence to identify trends in social media without understanding the implications of the more in-depth knowledge available there.

If we collect tweets and financial data associated with our target stock symbols, process the data using frameworks to deduce a tweet's sentiment score, and then compare the results to changes in the stock price, we can create a useful momentum marker for the share market.

The frameworks will classify tweets by polarity to resolve these concerns. A tweet's sentiment can be classified as positive, neutral, or negative, depending on its polarity. After tweets are categorized, a daily score is computed.

We are particularly interested in the following main research questions:

- What potential effects could tweets about a particular company have on the future trends of its share?

- Is it possible for the sentiment analysis results to be used to determine whether a stock's price is strong or weak in terms of momentum?

Hence, understanding whether sentiment can be used to predict the direction of trends for a share price is a crucial part of the research.

## 1.3 Methodology

As implied by the preceding sections, this study's methodological approach entails gathering data from tweets and then categorizing them based on their sentiment polarity. However, because the scope of the world's stock markets is immense, this document will rely on some of the most frequently occurring companies in the S&P 500 index. This decision was made for many reasons, including that the S&P 500, one of the world's most widely used and relevant indices, is recognized as the best measure of US big-cap equities, and its companies are popular on social media platforms.

The study was performed using the Python programming language. Additionally, Jupyter Notebook and the PyCharm IDE were used to implement the scripts created throughout the various modules.

The official Twitter API can be used to gather tweets. However, given the vast amount of data we seek, this has several limitations and is inappropriate for our case. The fact that there are few options makes finding Twitter data one of the study's biggest challenges. Although there were other options, such as Twint, a comprehensive and reliable Python web scraper, or Tweepy, a Python library for extracting tweets, we used a preprocessed dataset from the Kaggle website [13] to get around these constraints.

We used the top 5 index companies of the S&P 500, the hashtag "# STOCK," the cahshtag "$ STOCK," and a carefully curated and annotated dataset of tweets from April 9th to July 16th, 2020, to explain the relationship between stock price returns and sentiment reflected in tweets.

## 1.4   Document Structure

Five chapters are comprised in this document, each describing the steps required to achieve the objectives outlined in previous sections. Aside from Chapter 1, which contains some introductions, the rest of this dissertation is as described below.

The second chapter describes several concepts in the field of sentiment analysis and provides details on different levels and types of these concepts. The key components are its associated ideas, the application of Twitter in this setting, and the various algorithmic working methods, specifically machine learning methods. This chapter has described the methodology, information on data collection, dataset preparation for sentiment analysis, and a brief overview of the simulation setup.

The foundational theories and approaches that support the practice of sentiment analysis on stock market data are presented in Chapter 3. In addition, it provides many concepts, techniques, and details about two blocks of algorithm implementation and data acquisition.

Chapter 4 provides a detailed evaluation of the trials carried out on the selected five stocks. Furthermore, it also discusses the relationship between financial facts and sentimental bullishness via diagrams to clarify the study's findings.

Finally, overall conclusions and a summary of this dissertation, as well as possible future research paths for further development, are presented in Chapter 5.

# Chapter 2

# Related Work

We will discuss sentiment analysis's state of the art in this chapter. The background and its underlying ideas are presented in the beginning. Then we take a look at how to use Twitter as a valuable tool in sentiment analysis. After that, we take a quick look at various algorithmic strategies, specifically machine learning methods. Finally, we will discuss various approaches to Twitter data collection, including their advantages and disadvantages, and our chosen approach.

## 2.1  The Analogy Between Emotions and Sentiments

In daily life, the terms "emotion" and "sentiment" are frequently used interchangeably. Nevertheless, they are treated as two distinct ideas in psychology. The well-known psychological model by Watson and Tellegen depicts the two-dimensional model of emotions in Figure 2.1 [14]. Psychologically speaking, emotions play a key role in motivating behavior. Paul Ekman, a renowned psychologist, proposed six fundamental emotions: happiness, surprise, sadness, anger, fear, and disgust [15]. Later, other psychologists expanded these fundamentals by including excitement, contempt, pride, embarrassment, and shame. In contrast, sentiments describe mental states formed as a result of emotion. The definition emphasizes how sentiment expresses a person's thoughts that are influenced by their emotions. Sentiment, as opposed to emotion, is more expansive. The sentiment, which is exceptionally well organized, generally links actions to an emotion.

In summary, four factors can distinguish emotion from sentiment, as shown in Table 2.1. Furthermore, all emotions can be viewed as a synthesis of these four sentiment-influencing variables. Thus, analyzing opinions and identifying sentiments is critical.

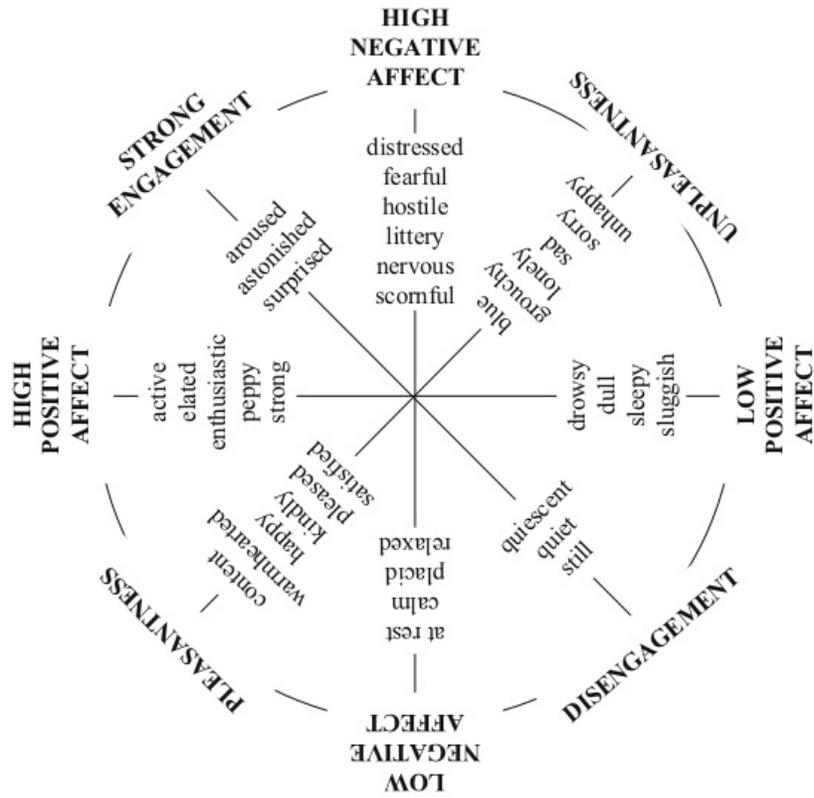

Figure 2.1 The two-dimensional structure of emotions
[16]

Table 2.1 The distinctions between emotion and sentiment

| | Definition | Connection | Dimension | Nature |
|---|---|---|---|---|
| **Emotion** | Psychological states | — | Psychological | Natural |
| **Sentiment** | Mental states | Expression of emotions | Social | Organized |

Table 2.1 The distinctions between emotion and sentiment
[16]

## 2.2 Sentiment Analysis

Sentiment describes how someone feels, either positively or negatively, about a topic. Opinion mining is another name for sentiment analysis. This is a natural language processing technique and is the study of a person's sentiments, opinions, and attitudes towards an entity such as a social event, product, or service. Text mining is the process of converting raw text with no structure to structured data in order to extract useful information. That is why studying algorithms that scan through the text and classify that text into different categories has received much attention.

Because of the popularity of social media platforms such as Twitter, Facebook, and YouTube, the number of comments containing individuals' opinions and information has increased dramatically; as a result, sentiment analysis has emerged as a hot and trending topic among academics and professionals. Sentiment analysis helps us make decisions about various subjects based on other people's opinions. Many companies use sentiment analysis to offer better products and services, run campaigns, and share promotional deals.

### 2.2.1 Sentiment Analysis Levels

As shown in Figure 2.2, sentiment analysis consists of three levels of granularity:

- Document Level

- Sentence Level

- Entity and Aspect Level

#### 2.2.1.1 Document-Level Sentiment Classification

This level identifies whether the concept of the entire document is positive or negative at this point by treating it as a whole. This approach presumes that a document discusses a single subject. For example, it determines whether a movie or book review is positive or negative overall.

#### 2.2.1.2 Sentence-Level Sentiment Classification

At this stage, it separates each sentence and classifies its opinion as positive, negative, or neutral. Unlike the document level, each document can have different opinions about multiple entities. Each sentence is classified into an objective or subjective type [18]. Subjective sentences are based on personal perspective and opinion, like "She has beautiful hair." Whereas the objective sentences do not consist of any people's opinions and are based on facts like the sentence "I have black hair."

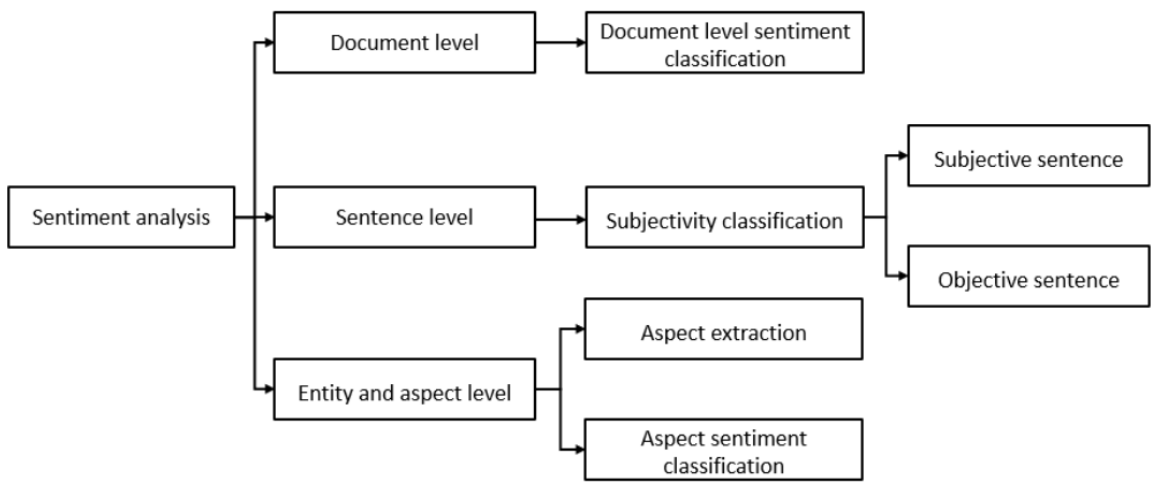

Figure 2.2 Sentiment Analysis Different Levels
[17]

### 2.2.1.3   Entity and Aspect Level Sentiment Classification

This level is also known as the feature-based level and, unlike document and sentence levels, approaches opinions directly. Its goal is to discover the opinions and determine their sentiments that could be positive, negative, or neutral about a subject or an entity.

## 2.2.2   Sentiment Classification Techniques

We got familiar with the sentiment analysis before and understand that the classification of opinions regarding entities is called sentiment analysis. Different strategies are used to collect opinions and associated sentiments; some are pretty sophisticated and time-consuming, and some are simple and fast.

As shown in Figure 2.3, we will look at them all:

- Lexicon-based approaches

- Machine learning approaches

- The Hybrid approach

### 2.2.2.1   Lexicon-Based Approaches

Lexicon-based techniques were among the first techniques used for sentiment analysis. Each word is given a polarity score in lexicon-based systems, and after going through the text, the system gives a sentiment score to that text. The advantage of this method is that it does not require data automation [20].

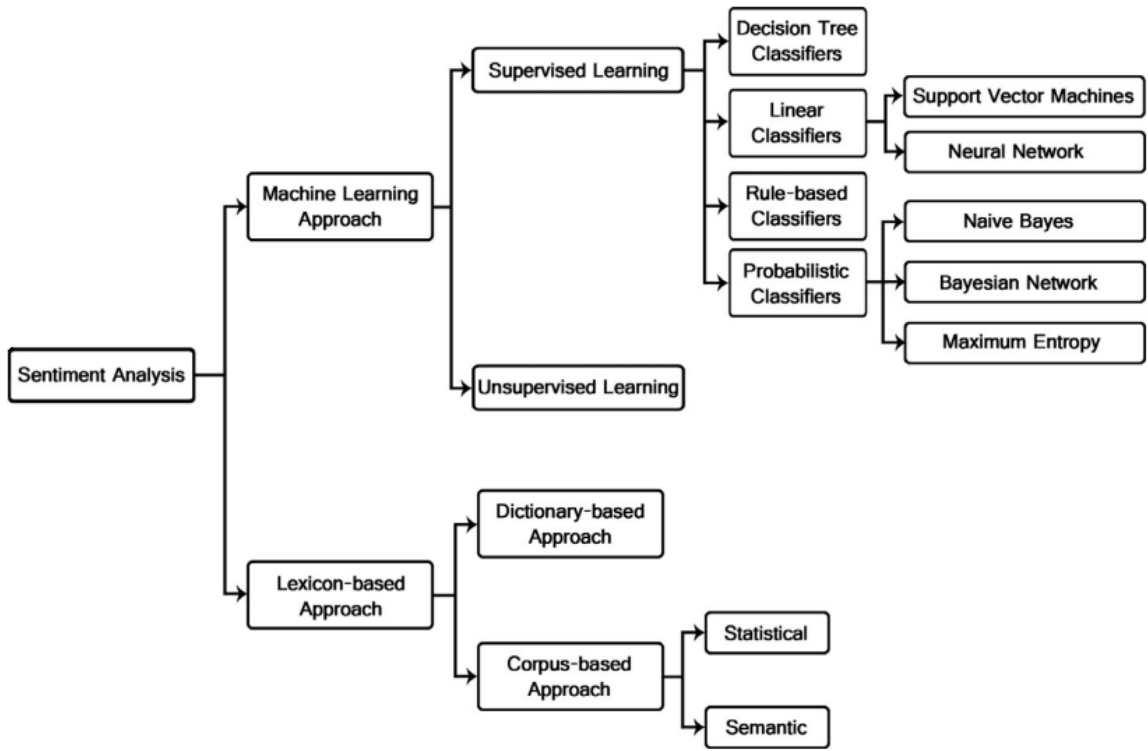

Figure 2.3 Different Sentiment Classification Techniques

[19]

The lexicon-based technique is divided into two sub techniques which are dictionary-based and corpus-based. Dictionary-based is done using a dictionary of terms like WordNet [21]. On the other hand, corpus-based does not need a predefined dictionary; instead, it uses statistical analysis approaches such as K-nearest neighbors.

### 2.2.2.2 Machine Learning Approaches

Machine learning is a branch of artificial intelligence that uses data as input to make better judgments rather than being explicitly instructed on what to do.

Machine learning approaches in sentiment analysis [22] can derive opinions and their sentiment towards an entity. For this purpose, various methods are employed, including support vector machines (SVM) and Naïve Bayes.

Generally, machine learning techniques are divided into two categories:

- Supervised learning

- Unsupervised learning

**Supervised Learning**    The goal of this technique is to predict outcomes in uncertain situations.

The model is first trained using data with a known output and label. Then, the system will learn from this data to make the best decision.

There are two types of supervised learning techniques:

**Regression**    The goal is to predict a real value based on different independent inputs. For example, determining a budget based on sales.

**Classification**    Placing various items in different categories. This is the approcehe that we use for sentiment analysis.

Serval algorithms are used to classify objects. For example, a support vector machine separates every object into two different classes, called binary classification. It tries to find a line to separate the members of the classes. Then it updates the values of the slope and parameters of that line based on each piece of data to find the best line for this classification.

**Unsupervised Learning**    Unsupervised learning differs from supervised learning since it does not base its decisions on a pre-trained model. Instead, it analyzes raw data to make the optimal judgments by attempting to extract patterns and similarities between the data without prior knowledge.

Unsupervised learning techniques are also divided into two sub categories:

**Clustering**    This Technique divides unlabeled data into groups with structures based on similarities and contrasts between the items. Clustering algorithms might be hierarchical, overlapping, or probabilistic.

K-means clustering is an example of clustering where the data is separated into K groups, and each set of data closest to each centroid is categorized under that category. Each group's size and level of detail can be altered by altering K, like in Figure 2.4.

**Association**    It is the process of identifying patterns and connections among many elements of a vast body of data.

### 2.2.2.3    The Hybrid Approach

The hybrid sentiment analysis approach uses both statistical and knowledge-based methodologies for polarity detection. As a result, it inherits both stability and excellent accuracy from the lexicon-based method and machine learning [24].

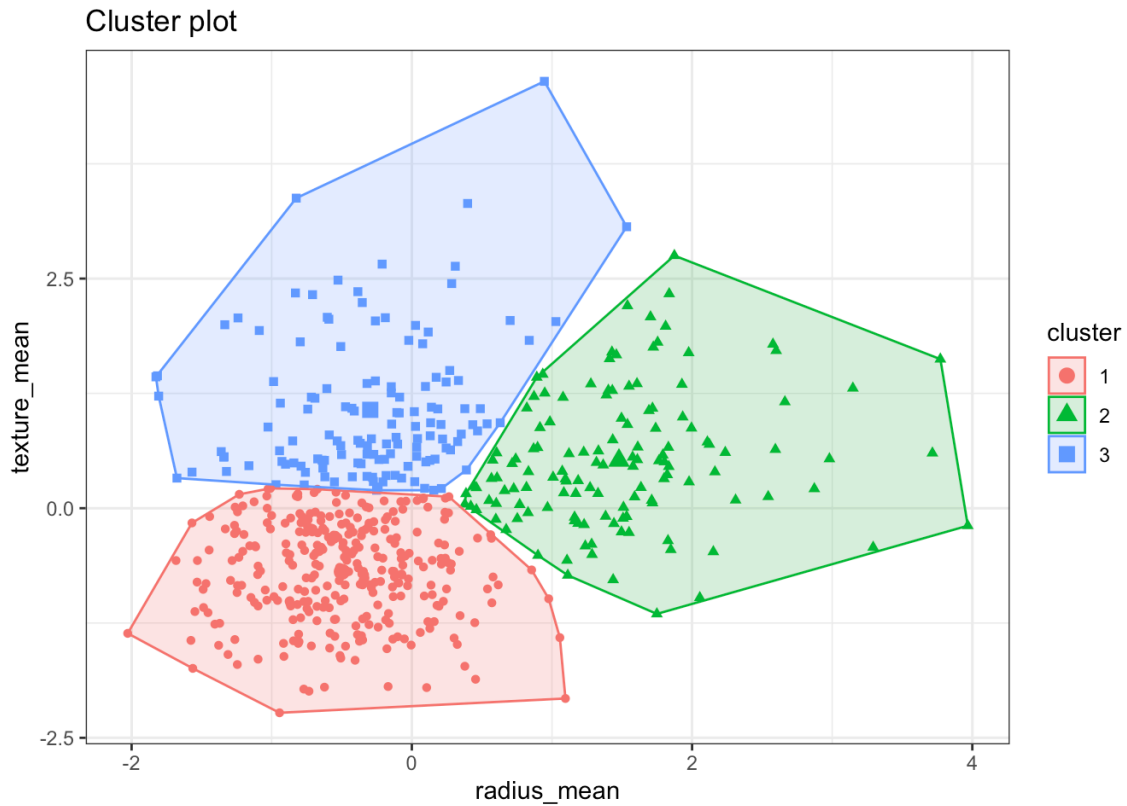

Figure 2.4 K-means' propensity to form clusters of the same size

[23]

### 2.2.3 Algorithmic Approaches

Numerous algorithms have been proposed for extracting sentiments regarding an entity or topic. Among the most popular algorithms for sentiment analysis are Naïve Bayes, Random Forest, and Long Short-Term Memory. We shall quickly review these algorithms in detail.

#### 2.2.3.1 Naïve Bayes Classifier

One of the most straightforward and least time-complex supervised algorithms for datasets containing big data with millions of records is Naïve Bayes. According to Naïve Bayes, there is no correlation between the incidence of one character in a class and the occurrence of another feature. Because of this, it is known as a "Naïve" algorithm.

The Bayes theorem, shown in Figure 2.6, on which the Naïve Bayes algorithm is based, enables us to calculate the posterior probability, for instance, $P(H|E)$, using $P(E|H)$, $P(H)$, and $P(E)$ where $P(H|E) = P(e_1|H) * P(e_2|H) * ... * P(e_n|H) * P(H)$.

To better understand how the Naïve Bayes algorithm works, consider the scenario of differentiating spam and genuine emails by scanning each email and extracting words and phrases. In order to determine if an email is spam or not, we create a table that contains the likelihood of discovering a word inside the message. For example, given that the email is spam, we compute the likelihood of detecting the term "offer." Additionally, we have to consider the possibility that the email was not a spam email. Remember that we used the training dataset to calculate these probabilities.

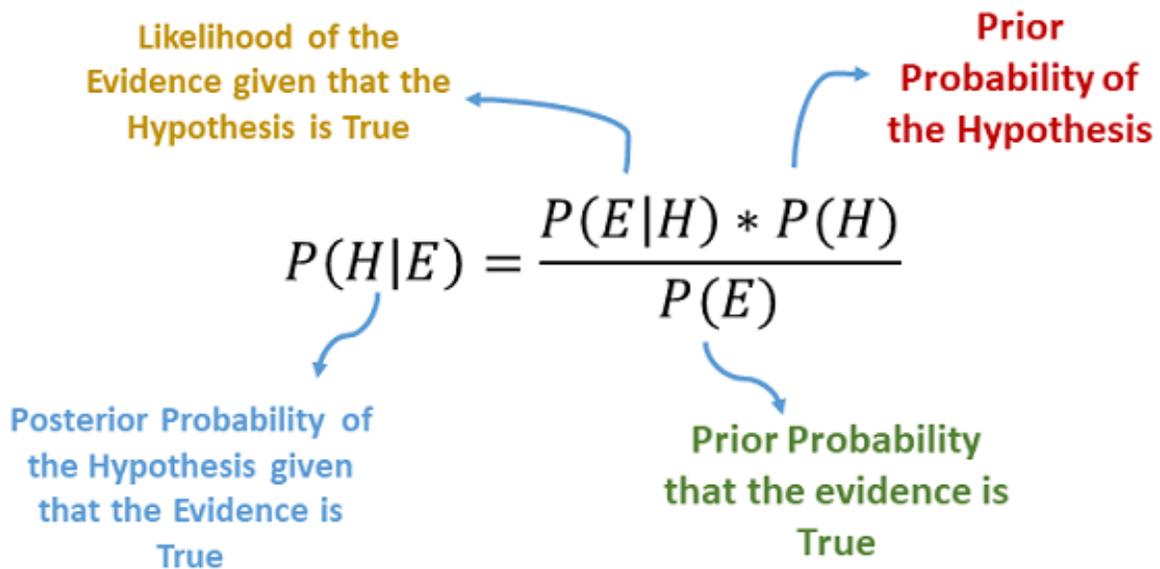

Figure 2.5 The assumption of conditional independence in the Naïve Bayes classifier
[25]

The followings are the advantages and disadvantages of the Naïve Bayes algorithm:

**Advantages of Naïve Bayes Classifier**

- It has a lower time complexity than other machine learning models.

- It performs better in classification problems.

- It needs less training data.

**Disadvantages of Naïve Bayes Classifier**

- The zero-frequency problem causes predictions to be ignored for categories with zero occurrences in the training data and zero probability.

- It is a terrible model for making numerical forecasts and estimates.

- This classifier tends to create assumptions that are mostly not valid in reality.

### 2.2.3.2 Random Forest Classifier

Random forest is another standard classification algorithm. The moniker "Random Forest" refers to a system that builds a lot of random decision trees and uses most of those predictions to determine the outcome. Each decision tree's features and data are chosen at random.

As seen in Figure 2.6, the dataset is divided into separate, entirely random decision trees, and when attempting to forecast an item, it employs a majority vote as the final prediction. In addition, it could be used to solve regression problems by calculating the mean or average of each decision tree prediction.

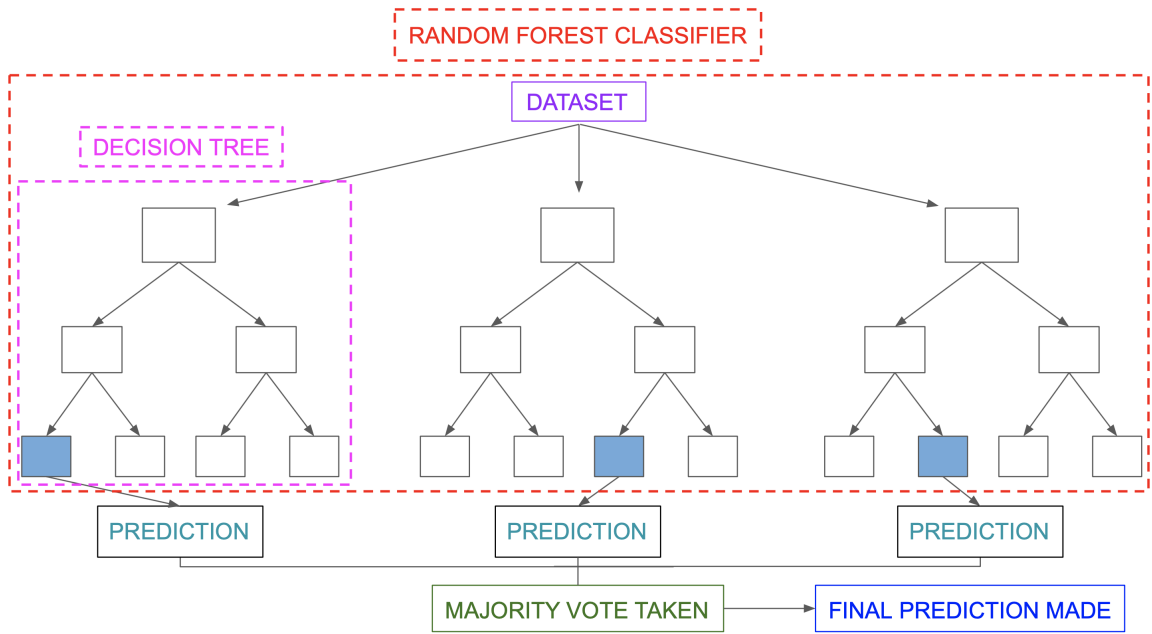

Figure 2.6 An example of how the Random Forest classifier functions
[26]

This algorithm has several advantages and disadvantages like other algorithms:

**Advantages of Random Forest Classifier**

- Making reasonable forecasts

- Working with enormous datasets

- Superior to decision trees in accuracy predictions

- Handling missing data better than other algorithms

- Being less problematic than a decision tree model in terms of overfitting

**Disadvantages of Random Forest Classifier**

- High time complexity

- High space complexity

### 2.2.3.3   Long Short Term Memory Model (LSTM)

Long short-term memory models are a kind of recurrent neural network that is very effective in handling a large amount of sequential data. New recurrent neural networks differ from traditional ones because they can take feedback from previous outputs to generate their current output. The main feature of LSTMs is storing information for an extended period of time, and they do that using a cell state. The cell states' primary role is to store information from the past efficiently [27].

In each iteration, the LSTM must decide:

- What information to add to the cell state

- Delete useless information from the state.

- Produce the output based on the input and report it to the cell.

The LSTM model will not be affected by outliers in making its predictions and can train its parameters. Figure 2.7 depicts a simple mechanism of this model.

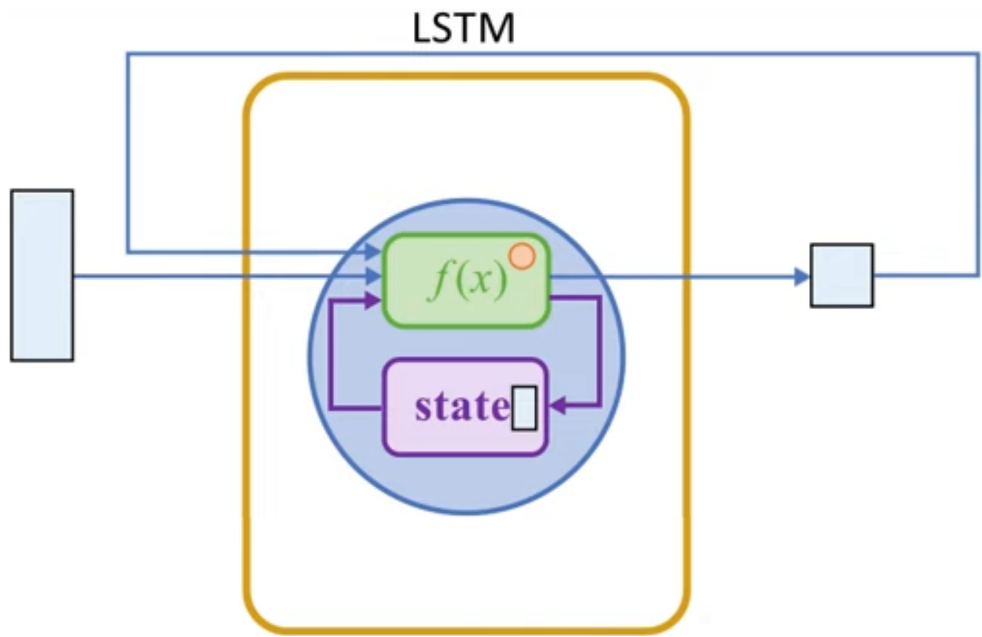

Figure 2.7 The architecture of the LSTM model
[28]

## 2.3  Applications of Sentiment Analysis

The number of people using social media is increasing every year. That is why millions of opinions are being shared through applications such as Twitter, so the amount of data containing opinions about various topics is growing rapidly. Therefore, employing sentiment analysis to extract opinions and their polarity has emerged as a trending study with numerous applications.

Here are some of the applications of sentimental analysis:

**Stock Market**    The stock market is crucial to every country's economy. Predicting how a stock will act in the upcoming future helps us make a better investment with our money. Sentimental analysis extracts opinions about stocks and tries to find the correlation between the opinions and the stock by gathering different datasets and processing them. With the help of sentimental analysis, we could try to predict a stock's trajectory after data collection and preparation.

**E-commerce**    Various e-commerce businesses use different types of feedback, such as reviews and comments about their products, to improve the quality of their products and customer service. Recommender systems also use sentiment analysis to analyze people's feelings toward entities and improve their recommendations. For example, Amazon has used sentiment analysis to improve the quality of the different services it provides.

So, the popularity of sentiment analysis is due to the great interest shown by these applications.

### 2.3.1 Twitter Sentiment Analysis

Twitter is one of the most used social media applications worldwide; launched in 2006, it has more than 400 million active users monthly. The main reason behind twitters creation was to allow its users to express their opinions about different topics in a tweet which is a text with less than 280 characters. As a result, Twitter users vary significantly from ordinary people to highly influential characters like Jeff Bezos and political figures like Barack Obama.

The main difference between Twitter and other social media applications is that everything you post on Twitter is available to the public even if they do not have Twitter; this feature makes sentiment analysis very effective on Twitter. Health care and politics are among the applications of sentimental analysis on Twitter.

**Politics** With the emergence of Twitter, many studies have been conducted to replace traditional polling methods like telephone poles and internet polls with Twitter sentiment analysis. Studies showed that algorithms like SVM classifier could help us predict the polls with high accuracy. Detecting sarcasm about a candidate is the most challenging part. In this case, positive sentiments are viewed negatively.

**Business** Many companies use Tweets to extract sentiment and feelings toward their product directly or indirectly. This helps companies to improve their product and offer better services.

## 2.4 Data Collection and Preparation

Data collection and Preparation is gathering raw data from multiple sources and coveting that into structured data to use for sentimental analysis. The main job of data preparation is to clean up into a single format.

In the case of Twitter, there are five steps to data preparation [29–31]:

1. Removing pictures and URLs

2. Removing usernames

3. Removing special characters: Here, punctuation marks must be removed. Another type of special character that is removed and replaced with space is emoticons. Even though it may seem like emoticons help discover a tweet's sentiment, they are too complex to interpret and be used in sentimental analysis.

4. Removing letters that contain no emotion, such as "a", and "the," which are also known as stop characters

5. Converting everything to lower case

### 2.4.1 Twitter Data Collection Methods

Several methods are available to collect tweets for use by sentiment analysis models. This study will examine the pros and cons of the different data collation methods such as Twitter API, Twint web scraper, Tweepy, and preprocessed datasets.

#### 2.4.1.1 Twitter API

We could use the official Twitter API for collecting data. The Twitter API comes in 3 subscriptions, standard, premium, and enterprise, where the standard is free to use. The main limitation of the Twitter API is that you can only fetch 3200 tweets from the past seven days, and it also needs an account to use. To overcome these limitations, we can subscribe to premium services, but they are not free. That is why we prefer to use libraries and tools such as Twint and Tweepy.

#### 2.4.1.2 Tweepy

Tweepy is a Python library used for collecting Twitter data. It uses the Twitter API for collecting and gathering tweets. Tweepy has complete documentation from beginner level to advanced. Moreover, it allows its users to post tweets and retweets and even has the ability to like tweets.

Despite this, it is a paid application, and its free version also has the same limitations as the Twitter API.

#### 2.4.1.3 Twint

Twint uses scraping to collect tweets and does not depend on the official Twitter API. With Twint, you could collect unlimited tweets with no limitations. Nevertheless, the documentation on Twint is incomplete and outdated compared to the recent updates on Twitter and cannot be used for posting or liking tweets.

The main differences between Twint and Tweepy are as follows:

- Tweepy uses the Twitter API to collect tweets, while Twint uses web scraping to gather tweets.

- Tweepy has the same limitations as the Twitter API in the number of tweets it can collect, while on the other hand, Twint has no limitations in the number of tweets it can gather.

- Due to its excellent documentation, Tweepy is much easier to learn and use. On the other hand, in the case of Twint, because of its weak documentation, it can be a hassle to learn and use its features.

- Twint is limited to only collecting tweets and gathering information, while Tweepy can be used for posting and liking tweets.

#### 2.4.1.4 Preprocessed Datasets

Using preprocessed datasets is the alternative to the Twitter API and other third-party applications. Employing preprocessed datasets removes the trouble of trying to collect the tweets but has less flexibility than tweet collecting tools.

For example, the Kaggle website offers many preprocessed datasets that can be used for different studies.

We have chosen this method for our data and used a semi-preprocessed dataset from the Kaggle website.

# Chapter 3

# Methodological Approaches and Technical Frameworks

## 3.1 Data

In this section, we explain the details of the datasets that we have used, the purpose of those specific datasets, the operations that we performed to make data ready for use, and the procedure of labeling the data.

### 3.1.1 Twitter Data Collection

We used two datasets collected from twitter.

The first dataset consists of 1.6 million labeled records, containing tweets that are not necessarily related to a specific topic.

We obtained this dataset from deepnote.com to train our natural language processing models. Half of the records are labeled as 0 and half as 4. The labels represent the impression of the tweets. Label 0 indicates a negative vibe and 4 represents a positive attitude.

After building and testing the classifier models, We used the models to label a second dataset from the Kaggle website. This dataset consists of around 90,000 records of tweets relevant to stock markets and three columns comprising an id, created date, and full text of the tweets. This data was collected during a period of three months, from April 9th to July 16th, 2020.

The tweets from this dataset contain the names of more than 13000 stocks. We extracted the data related to only five stock markets: Facebook (which has recently changed to Meta), Microsoft, Tesla, Apple, and Amazon, which are the most popular and highly traded ones.

```
top5_dataframe_noLable = data[data['full_text'].str.contains('|'.join(
    top5_tickers))]
```

Listing 1 Extracting data of the top 5 stocks

Then we preprocessed the data and created a separate data frame for each stock:

```
AAPL_DF = stockDF[stockDF['full_text'].str.contains('aapl')]
AMZN_DF = stockDF[stockDF['full_text'].str.contains('amzn')]
FB_DF = stockDF[stockDF['full_text'].str.contains('fb')]
MSFT_DF = stockDF[stockDF['full_text'].str.contains('msft')]
TSLA_DF = stockDF[stockDF['full_text'].str.contains('tsla')]
```

Listing 2 Separating the data of each stock

### 3.1.2 Financial Data Collection

We collected the market data from the Yahoo Finance API for five stocks: Meta, Microsoft, Tesla, Apple, and Amazon.

Yahoo Finance API is a free API with an impressive amount of information and news on different markets. Extracting data from it is also pretty simple.

```
import yfinance as yf
start = AAPL_DF['Date'].iloc[0]
end = AAPL_DF['Date'].iloc[-1]
apple_data = yf.download("AAPL", start=start, end=end)
meta_data = yf.download("META", start=start, end=end)
micro_data = yf.download("MSFT", start=start, end=end)
tesla_data = yf.download("TSLA", start=start, end=end)
amazon_data = yf.download("AMZN", start=start, end=end)
```

Listing 3 Downloading data from Yahoo Finance

This data contains six attributes: Open, High, Low, Close, Adj Close, and Volume. We only Close, to calculate the amount of *Return* and check the correlation with twitter data, respectively.

### 3.1.3 Data Preprocessing

Data needs preprocessing, both before being used to build a supervised learning model, and before being labeled by that model.

We explain what we did to preprocess both twitter datasets.

The first important prerequisite to text preprocessing is to clean the text of the tweets from pronouns and stative verbs, URLs, emojis, usernames, numbers, and punctuations. We put pronouns and stative verbs in a list called *stopwordslist* and cleaned the text of the tweets from them:

```
  STOPWORDS = set(stopwordlist)
2 def cleaning_stopwords(text):
  return " ".join([word for word in str(text).split() if word not in
   STOPWORDS])
```

Listing 4 Cleaning data from redundant words

We removed punctuations:

```
1 english_punctuations = string.punctuation
  punctuations_list = english_punctuations
3 def cleaning_punctuations(text):
  translator = str.maketrans('', '', punctuations_list)
5 return text.translate(translator)
```

Listing 5 Removing punctuations

Removed repeating characters, for example converts *hiiiii* to *hi*:

```
1 def cleaning_repeating_char(text):
  return re.sub(r'(.)1+', r'1', text)
```

Listing 6 Cleaning repeating characters

Removed URLs and numbers:

```
  def cleaning_URLs(data):
2 return re.sub('((www.[^s]+)|(https?://[^s]+))',' ',data)
  def cleaning_numbers(data):
4 return re.sub(r'[\W_]+',' ',data)
```

The next step is to tokenize the words in the text so that it will be easier to go through and process. We changed the text of each tweet into a vector of words.

```
  tokenizer = RegexpTokenizer(r'\w+')
```

We also stem the words. It means converting the different words with the same root to their single root. e.g., the root of all the words payment, pay, and paying is the word pay. Having the stem word instead of these different forms helps the procedure of processing and learning easier while maintaining the main meaning and context.

```
  import nltk
  st = nltk.PorterStemmer()
3 def stemming_on_text(data):
  text = [st.stem(word) for word in data]
```

```
5   return data
```

Listing 7 Stemming

Another necessity is to *lemmatize* the words, meaning we omit all different shapes of one word using the nltk built-in function *WordNetLemmatizer()*.

```
1   nltk.download('wordnet')
    lm = nltk.WordNetLemmatizer()
3   def lemmatizer_on_text(data):
    text = [lm.lemmatize(word) for word in data]
5   return data
```

Listing 8 Lemmatizing

We applied all the created functions using lambda expression sing the following syntax:

```
data = data.apply(lambda x: function(x))
```

Listing 9 Applying the preprocessing functions

And further did a one-hot encoding on each word, converting each word to a vector to be used for processing.

One-hot encoding (Figure 3.1) represents each word with a vector containing N values, N being the total number of words in the database. All the values in this vector are 0, except the one for which the index belongs to the target word. As the number of words in the database is enormous, we omitted the least repeated words and kept the most repeated ones to make processing workable for an extensive database.

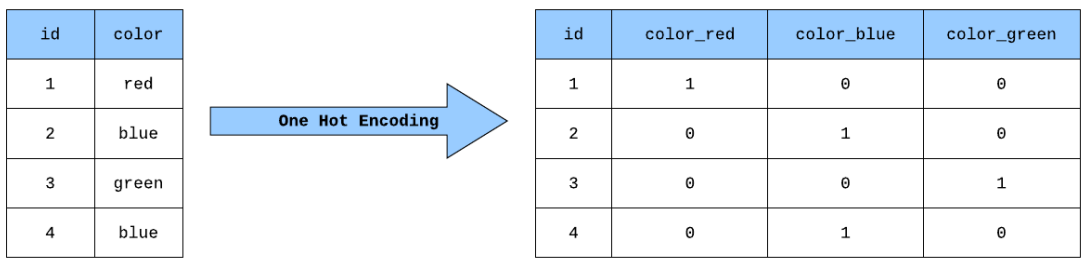

Figure 3.1 An example of one-hot encoding
[32]

Once preprocessing is done, we are good to go to fit our model into the data. For this purpose, we split the data into train and test sections and built our classifier models.

### 3.1.4 Labelling

We labelled the data extracted from our second dataset, by the classifier models built using the first dataset. As mentioned before, the labels of our training data were 0 and 4. For the sake of simplicity, we replaced label 4 with 1.

Labelling using the Random Forest model:

```
rfPred = np.where(rfPred==4,1,rfPred)
stockDF = dup_top5_noLable.copy()
stockDF['Label_rf']=rfPred
```

Listing 10 Labelling using Random Forest model

Labelling using the Naïve Bayes model:

```
nbPred = BNBmodel.predict(X_test)
nbPred = np.where(nbPred==4,1,nbPred)
stockDF['Label_nb']=nbPred
```

Listing 11 Labelling using Naïve Bayes model

Labelling using the LSTM algorithm:

```
test_sequences = tokenizer.texts_to_sequences(dup_top5_noLable['
  full_text'])
test_padded = pad_sequences(test_sequences, padding='post', maxlen=
  max_length)
pred_labels = []
prediction = model.predict(test_padded)
for i in prediction:
if i >= 0.5:
pred_labels.append(1)
else:
pred_labels.append(0)
pred_labels
stockDF['Label_lstm'] = pred_labels
```

Listing 12 Labelling using LSTM model

## 3.2 Algorithmic Implementation

In this section, we explain the details of building the classifier models, including the code and parameters, hyper-parameter tuning considerations, the structure of the implemented models, checking the accuracy, and the concepts that are necessary to comprehend the code.

### 3.2.1 Random Forest Classifier

We used random forest classifier model from *sklearn*. There are two parameters passed to the classifier model that we wish to explain: the number of estimators and the maximum depth.

The number of estimators represents the number of decision trees we wish to generate. A higher number of trees guarantees a more accurate prediction, but increases the amount of calculations and slows down the processing in return. Therefore, the optimal choice for this parameter is the largest that our processor can handle.

Max depth represents the maximum depth of each tree. If we do not set a maximum depth, then the algorithm continues to generate leaves until it runs out of samples, or all of the leaves are pure.

There are also other critical parameters that can be adjusted such as *max_features* and *min_sample_leaf* which modify the performance and speed inseparably like *n_estimators* did. There are also some parameters that can directly modify the speed of training including *n_jobs*, *random_state*, and *oob_score*.

```
model = RandomForestClassifier(n_estimators=200,max_depth=60)
model.fit(X_train,y_train)
predictions = model.predict(X_test)
```

Listing 13 Buliding Random Forest model

We further use a confusion matrix to observe the result. A confusion matrix is a table containing four cells: TP (True Positive), TN (True Negative), FP (False Positive), and FN (False Negative).

- TP indicates the number of accurately predicted values that are positive.

- TN shows the number of values that are predicted as negative and are actually negative.

- FP keeps count of the negative values that are predicted as positive.

- FN represents the number of positive values that are falsely predicted as negative.

```
from sklearn.metrics import confusion_matrix
confusion_matrix(y_test,predictions)
```

Listing 14 Showing Confusion Matrix

And checked the accuracy score, representing the percentage of records the label of which our model has predicted accurately.

```
from sklearn.metrics import accuracy_score
accuracy_score(y_test,predictions)*100
```

<div align="center">Listing 15 Accuracy of predictions of the Random Forest model</div>

### 3.2.2   Naïve Bayes (Bernoulli) Classifier

The implementation can be simply done by calling the Bernoulli model's built-in fit function on the train and test data and predicting and checking the accuracy.

The most critical optimization parameters of this algorithm are *alpha*, *fit_prior*, and *class_prior*.

```
from sklearn.Naïve_bayes import BernoulliNB
BNBmodel = BernoulliNB()
BNBmodel.fit(X_train, y_train)
y_pred1 = BNBmodel.predict(X_test)
accuracy_score(y_test,y_pred1)*100
```

<div align="center">Listing 16 Building Naïve Bayes model</div>

### 3.2.3   Long short-term memory Model(LSTM)

We built a sequential 4-layer model, starting with 200 features and ending with 1. The first layer is a neural network with 200 starting nodes, whose values are passed into the middle nodes to get the values of the 100 ending nodes, whose values are returned as the output. Obviously, we need the model to return one value only, and that is the predicted value.

The embedding layer works like a lookup table that helps embed each word in the vector of the total words existing in the dataset. It specifies an integer value for each word and maps it to the dense vectors. The embedding layer helps embed high-dimensional data into lower-dimension space.

A bidirectional layer is a combination of two RNNs, one of which data passes through beginning from the start of the data sequence, and the other through which data moves in the opposite direction. The need for this kind of traversal can be perceived by the fact that in text processing, the meaning of a word is not only dependent on the previous words but also on the following ones.

We use activation functions in the dense layers.

```
import keras
model = keras.Sequential([
keras.layers.Embedding(vocab_size, embedding_dim, input_length=
  max_length),
```

```
    keras.layers.Bidirectional(keras.layers.LSTM(64)),
5   keras.layers.Dense(24, activation='relu'),
    keras.layers.Dense(1, activation='sigmoid')
7   ])
```

<div align="center">Listing 17 Sequential LSTM layers</div>

Activation functions are necessary for a deep model to learn appropriately as they fix the neuron values to prevent them from becoming so big or small that they become useless. Popular activation functions are Relu, sigmoid, and tangent.

ReLU function is considered to be the most reliable activation function to be used in between the layers. It returns the input itself unless the input is negative, in which case it returns zero. The sigmoid and tangent activation functions are used in classification problems.

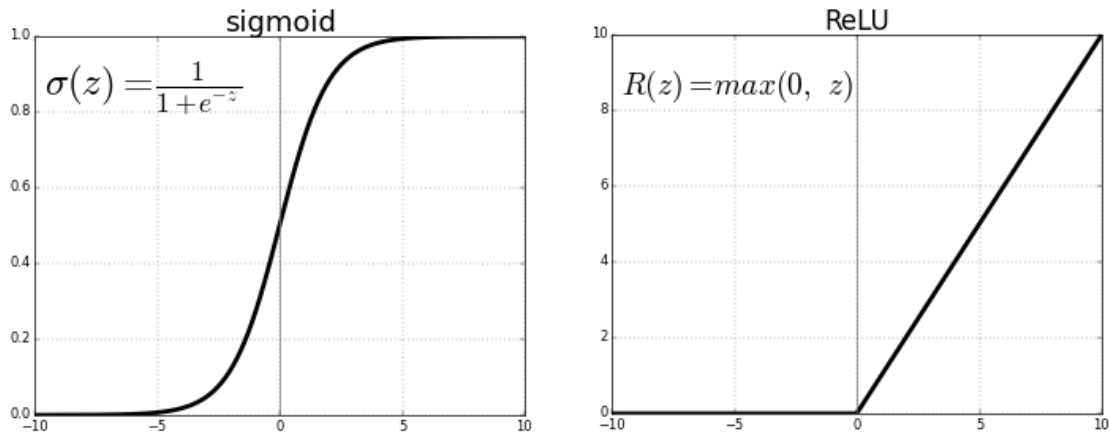

<div align="center">Figure 3.2 Sigmoid and ReLU Functions

[33]</div>

In order to proceed to the explanation of the code, we need to define the loss function and optimizer.

The loss function compares the prediction output of the model to the expected values, which are the actual values, and returns a number, or loss, for each observation. The model is trained best when the amount of loss is minimized.

The most popular loss functions are mean squared error (MSE), categorical cross-entropy, and binary cross-entropy. Each is used for a different kind of problem.

MSE works best for regression problems and cross-entropy for classification ones. As we are expecting our input to have only two values, we use binary cross-entropy.

The optimizer is the algorithm used to update the weights of the neurons using the gradient of the loss function. GSD and Adam are two of the most stable optimizers. The only parameter of the optimizer that may require adjustments is the learning rate. If the learning rate is too small, it searches for the minima of the loss function in very tiny steps, resulting in a super-slow process of learning. Otherwise, if the learning rate is too large, it may learn quickly initially but may not find the minima due to the large steps it takes while searching, resulting in an unstable training process.

```
model.compile(loss='binary_crossentropy',
optimizer='adam',
metrics=['accuracy'])
model.summary()
```

Listing 18 Compiling the LSTM model

The batch size specifies the number of observations passed to the network at each training step. The number of epochs assesses the number of times that the full training data is shown to the neural network.

```
num_epochs = 2
history = model.fit(train_padded, lstm_y_train,
epochs=num_epochs, verbose=1,
validation_split=0.1)
```

Listing 19 Fitting the LSTM model to the data

Setting the *verbose* attribute results in seeing the progress bar while running, and *validation_split* represents the fraction of data used for validation to the total amount of training data.

Eventually, we tested the accuracy of the model along with replacing the label 4 with 1.

```
prediction = model.predict(test_padded)
lstm_y_test = np.where(lstm_y_test==4,1,lstm_y_test)
pred_labels = []
for i in prediction:
if i >= 0.5:
pred_labels.append(1)
else:
pred_labels.append(0)
print("Accuracy of prediction on test set : ", accuracy_score(
  lstm_y_test,pred_labels))
```

Listing 20 Checking the accuracy of LSTM model

# Chapter 4

# Performance Evaluation

## 4.1 A Diagrammatic Structure for Performance Evaluation

We defined and calculated a variable *bullishness* ($B_t$) for each day, extracted from Twitter data features [34], as:

$$B_t = \ln(\frac{1 + M_t^{Positive}}{1 + M_t^{Negative}})$$

```python
def bullishness(data):
    newData = pd.DataFrame(columns=['Date','lstm','rf','nb'])
    newData.Date = data.created_at.unique()
    grouped = data.groupby(data.created_at)
    for date in newData.Date:
        dfTemp = grouped.get_group(date)
        lstmPos = (dfTemp.Label_lstm==1).sum()
        lstmNeg = (dfTemp.Label_lstm==0).sum()
        newData.loc[newData.Date == dfTemp.created_at.values[0],"lstm"]=math.
          log((1+lstmPos)/(1+lstmNeg))
        rfPos = (dfTemp.Label_rf==1).sum()
        rfNeg = (dfTemp.Label_rf==0).sum()
        newData.loc[newData.Date == dfTemp.created_at.values[0],"rf"]=math.log
          ((1+rfPos)/(1+rfNeg))
        nbPos = (dfTemp.Label_nb==1).sum()
        nbNeg = (dfTemp.Label_nb==0).sum()
        newData.loc[newData.Date == dfTemp.created_at.values[0],"nb"]=math.log
          ((1+nbPos)/(1+nbNeg))
    return newData
```

Listing 21 Calculating the Bullishness

And extracted a variable *Return* from the *Close* feature of financial data.

The Opening value of a stock is the value for which the stock is traded at the beginning of the day, and the Closing value is the price of a stock that has been agreed upon after all transactions of that day, or simply the price for the last transaction of the day for that stock.

To see how much the value of a stock has grown or dropped in one day relative to the day before, we calculate a variable *Return* for each day $t$ [34].

$$Return_t = (\ln Close_t - \ln Close_{t-1}) \times 100$$

```
  def calc_Return(data):
2 newData = pd.DataFrame(columns=['Date','Values'])
  for i in range(1,len(data.index)):
4 newData.loc[i-1] = [data.index[i],(math.log(data.Close.iloc[i])-math.
    log(data.Close.iloc[i-1]))*100]
  return newData
```

Listing 22 Calculating the Return value

Finally, we analyzed the correlation between the financial feature "Return" and the Twitter data feature "Bullishness".

## 4.2    Detailed evaluation of the results

In this section, we will depict the analysis and look for correlations between the financial and the Twitter data.

### 4.2.1    Evaluation of The Apple Stock ($AAPL)

We visualized the difference between the calculated bullishness and the output of each classification model for the period of three months.

```
1 AAPL_B = bullishness(AAPL_DF)
  import matplotlib.pyplot as plt
3 plt.plot(AAPL_B.Date,AAPL_B.lstm)
  plt.plot(AAPL_B.Date,AAPL_B.nb)
5 plt.plot(AAPL_B.Date,AAPL_B.rf)
  plt.legend(["LSTM","Naïve base","Random forest"])
```

Listing 23 Plotting three predictions of one stock

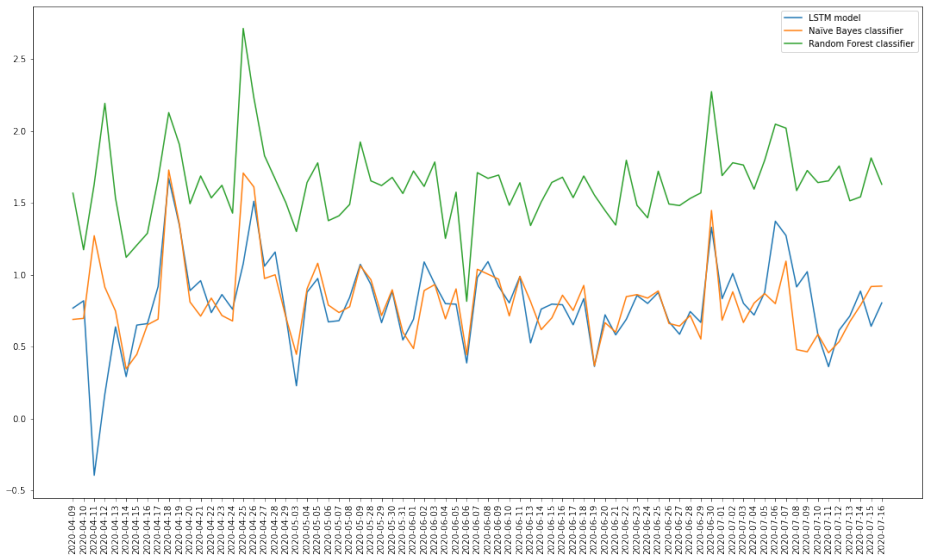

Figure 4.1 The Graph of AAPL Bullishness

As we can see in Figure 4.1, Naïve Bayes and LSTM models have pretty similar results, but far from the predicted bullishness of Random Forest model. If we look more precisely, we can recognize a similar pattern for all classifiers for most of the time intervals.

Here is the diagram of calculated return on the data extracted from Yahoo Finance statistics (graphical results are available in Figure **??**):

```
AAPL_R = calc_Return(apple_data)
plt.plot(AAPL_R.Date,AAPL_R.Values)
plt.show
```

Listing 24 Plotting the amount of Return of one stock

Then we merged the financial and Twitter results into one data frame using *FinRes*:

```
import time
import datetime
def finalRes(return_data,bullishness_data):
fin = pd.DataFrame(columns=['Date','lstm','rf','nb','Values'])
l =[]
for i in range(1,len(return_data.index)):
for j in range(0,len(bullishness_data.index)):
if datetime.datetime.strptime(bullishness_data.Date.iloc[j],"%Y-%m-%d" )>=return_data.Date.iloc[i]:
l.append([bullishness_data.Date.iloc[j-1],bullishness_data.lstm.iloc[j -1],bullishness_data.rf.iloc[j-1],bullishness_data.nb.iloc[j-1], return_data.Values.iloc[i]])
```

```
       break
11  fin = pd.DataFrame(l,columns=['Date','lstm','rf','nb','Values'])
    return fin
13  res = finalRes(AAPL_R,AAPL_B)
```

Listing 25 Merging the prediction and Return value into one dataset

We further used the merged data to compare each classifier output to the real market events by visualization.

```
1  def drawAllForStock(data,str):
   bull = bullishness(data)
3  start = bull['Date'].iloc[0]
   end = bull['Date'].iloc[-1]
5  stock_data = yf.download(str, start=start, end=end)
   stock_data.drop(apple_data.columns.difference(['Close']), 1, inplace=
     True)
7  ret = calc_Return(stock_data)
   res = finalRes(ret,bull)
9  drawPlot(res.Date,res.lstm,res.Values,"lstm model")
   drawPlot(res.Date,res.rf,res.Values,"Random forest model")
11 drawPlot(res.Date,res.nb,res.Values,"Naïve Bayes model")
```

Listing 26 Draw the comparison plots of Return and all predictions

And drew and saved the diagrams in one block:

```
1  drawAllForStock(AAPL_DF,'AAPL')
   figs = [plt.figure(n) for n in plt.get_fignums()]
3  i=0
   for fig in figs:
5  fig.savefig(f"apple_RB{i}.png", dpi=300)
   i = i+1
```

Listing 27 Drawing and saving the open figures

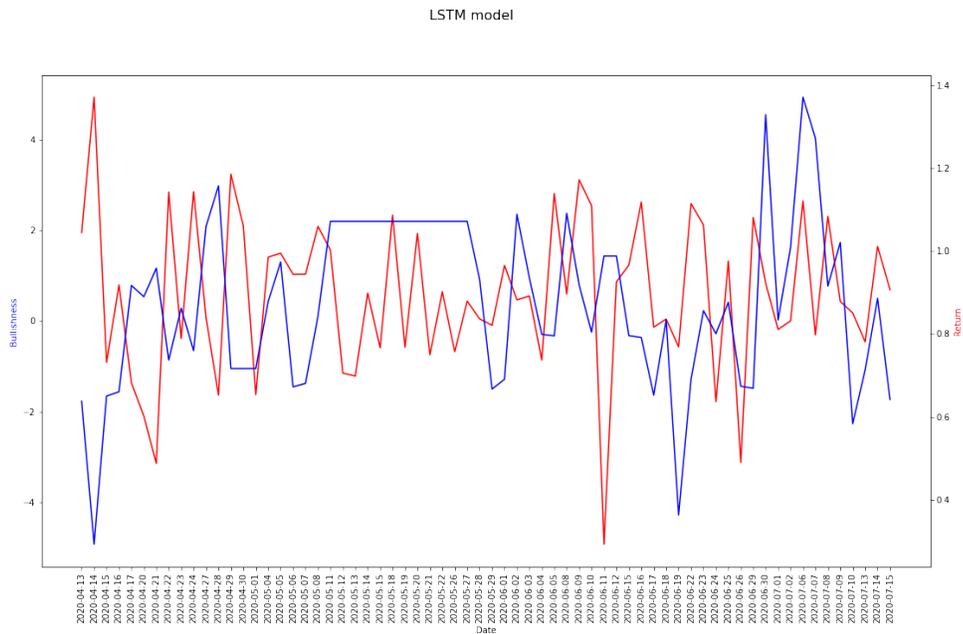

Figure 4.2 The Graph of the LSTM Model for AAPL

Considering Figure 4.2, for most days, the positivity or negativity of the Return approves of the predicted bullishness of the market, and there are close predictions on some days, such as June 25th to July 2nd. The gap between the prediction and the actual return after this date is quite understandable. As the stock has been growing rapidly, people would expect it to continue likewise, resulting in many positive comments. Nevertheless, the amount of the Return has dropped unexpectedly. The predictions are also pretty close after this gap to July 15th.

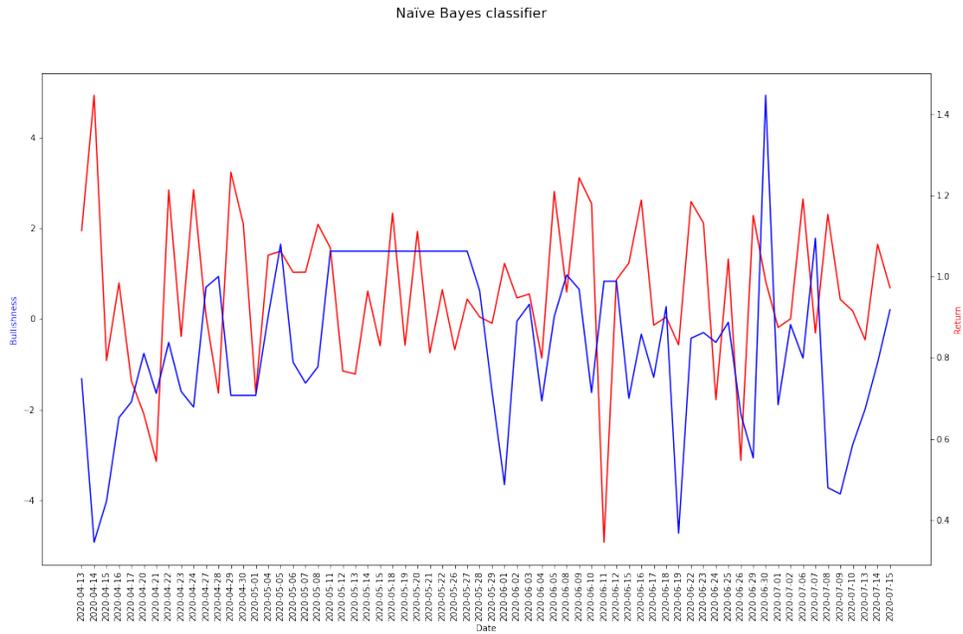

Figure 4.3 The Graph of the Naïve Bayes Classifier for AAPL

Figure 4.3 depicts that the Naïve Bayes classifier has probably predicted the worst. Though the predictions of this model are pretty similar to the Naïve Bayes', the concurrent gaps are larger for this classifier. One interesting point to take into consideration is that on some days, the amount of Return is pretty close to the prediction for tomorrow, meaning that there has been a closer relationship between the positivity of people and the amount of Return on the same day than the day after.

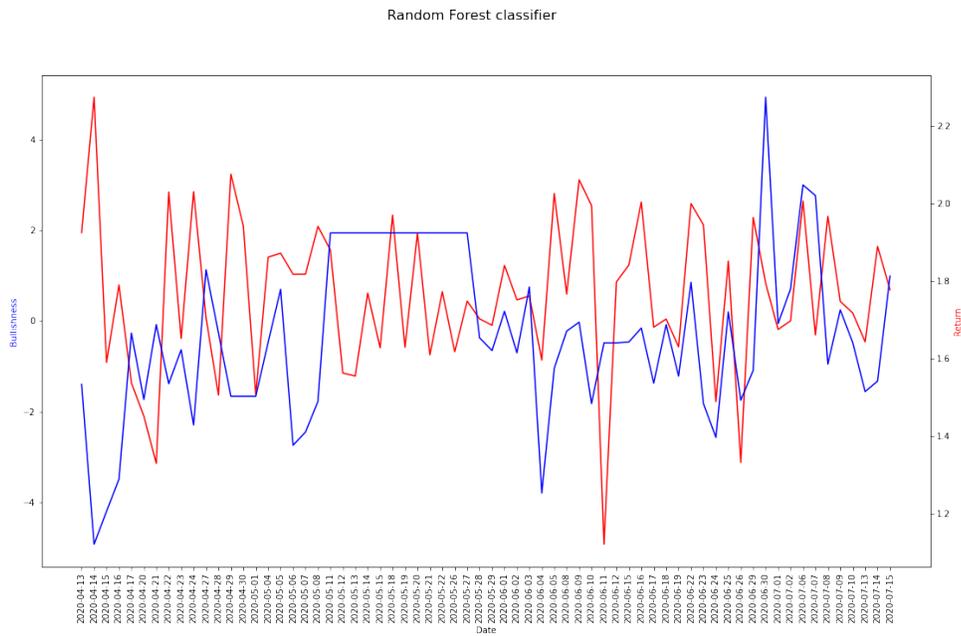

Figure 4.4 The Graph of the Random Forest Classifier for AAPL

Random Forest has predicted the most satisfactory, especially from June 18th on. (Figure 4.4)

## 4.2.2    Evaluation of the Facebook Stock ($FB)

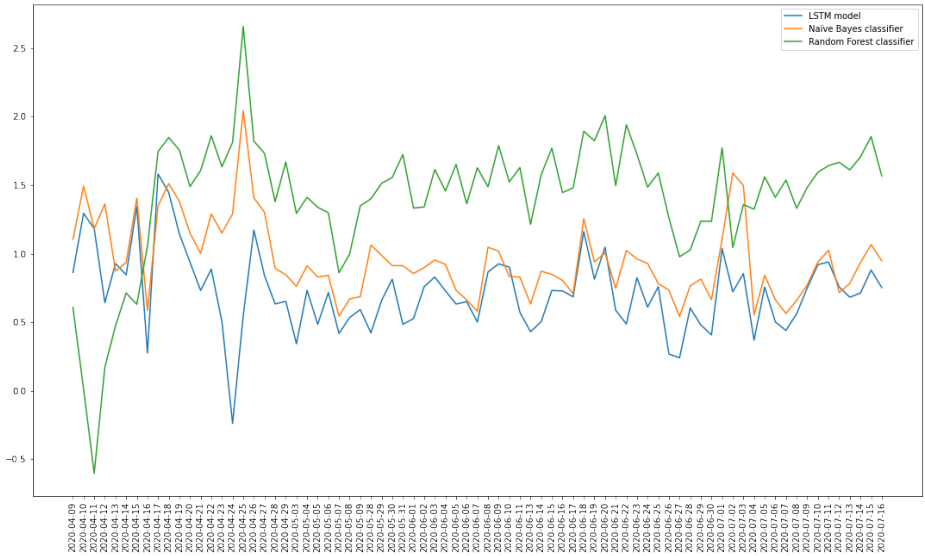

Figure 4.5 The Graph of FB Bullishness

As shown in Figure 4.5, like the Apple stock, Naïve Bayes and LSTM outputs are more similar compared to Random Forest. We can observe some similarities between the patterns.

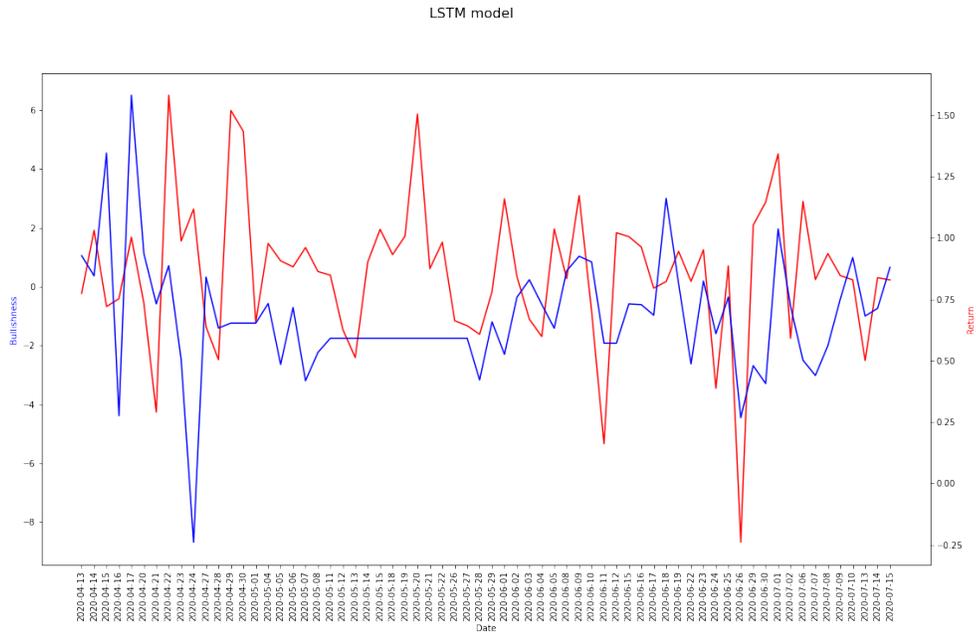

Figure 4.6 The Graph of the LSTM Model for FB

In Figure 4.6, we can see that there are some lucky days that predictions are close, but in most cases, the predictions are quite unreliable. The closest shot time interval is similar to that of Apple's, from June 23th to 29th.

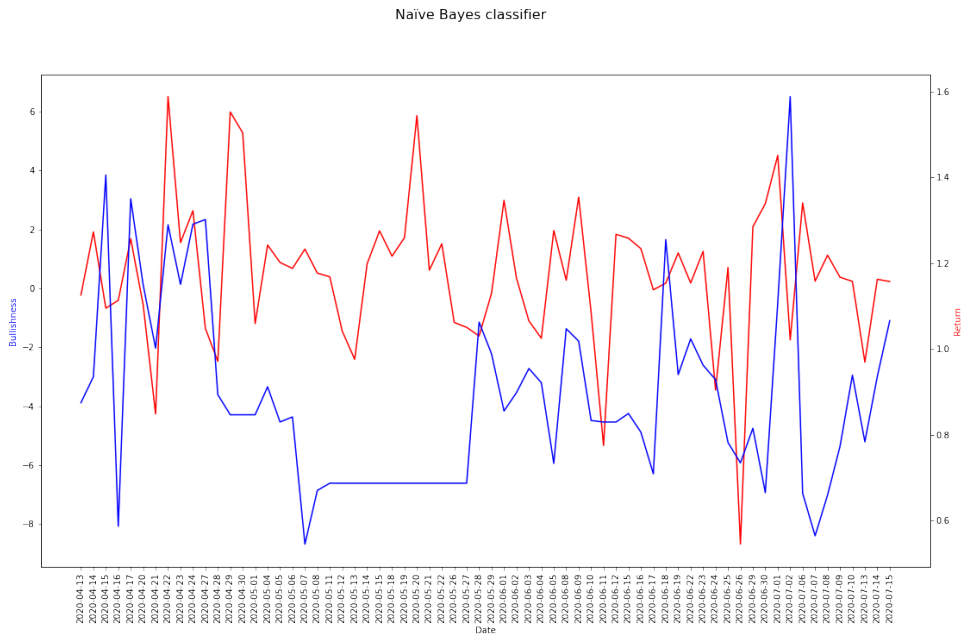

Figure 4.7 The Graph of the Naïve Bayes Classifier for FB

Naïve Bayes' predictions (Figure 4.7) are slightly closer than those of LSTM. The results are the most satisfactory from May 1st to 5th and June 24th to July 7th.

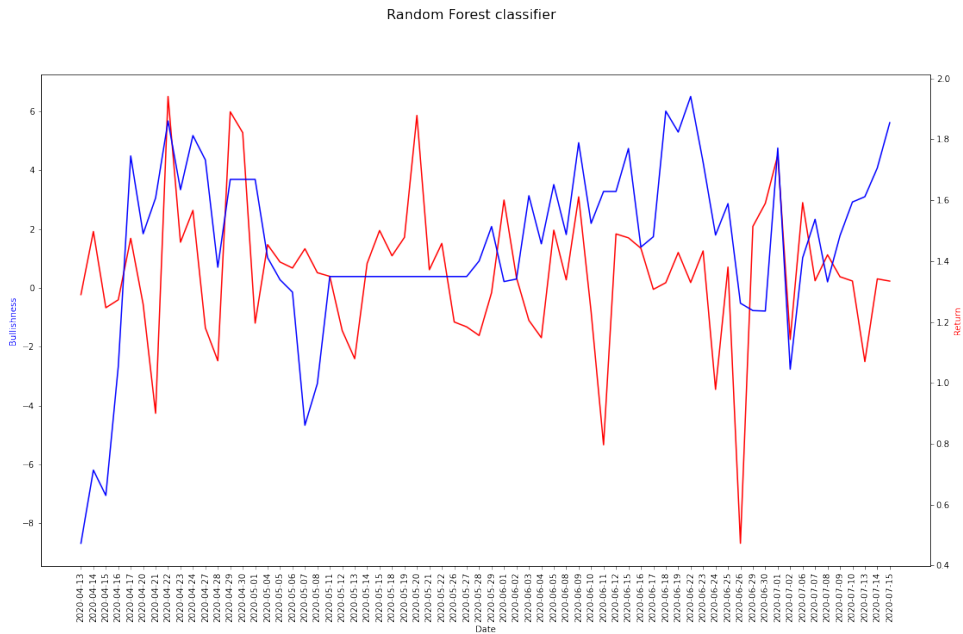

Figure 4.8 The Graph of the Random Forest Classifier for FB

As it is visible in Figure 4.8, many of the predictions of Random Forest are far away from the actual stock Returns, and a few are pretty close, making it difficult to evaluate the job. The closest predictions belong to the first days of July, in which the predictions of the other classifiers, even the ones belonging to the other stocks, have also been close.

### 4.2.3   Evaluation of the Microsoft Stock ($MSFT)

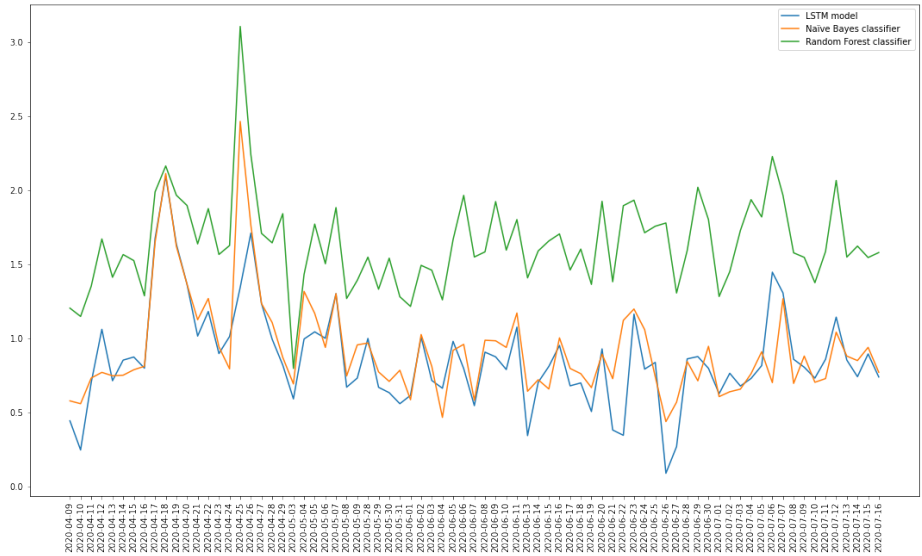

Figure 4.9 The Graph of MSFT Bullishness

Naïve Bayes and LSTM outputs are more similar compared with Random Forest (Figure 4.9). Many similarities between the patterns can be observed in all three. All classifiers agree on the peak points.

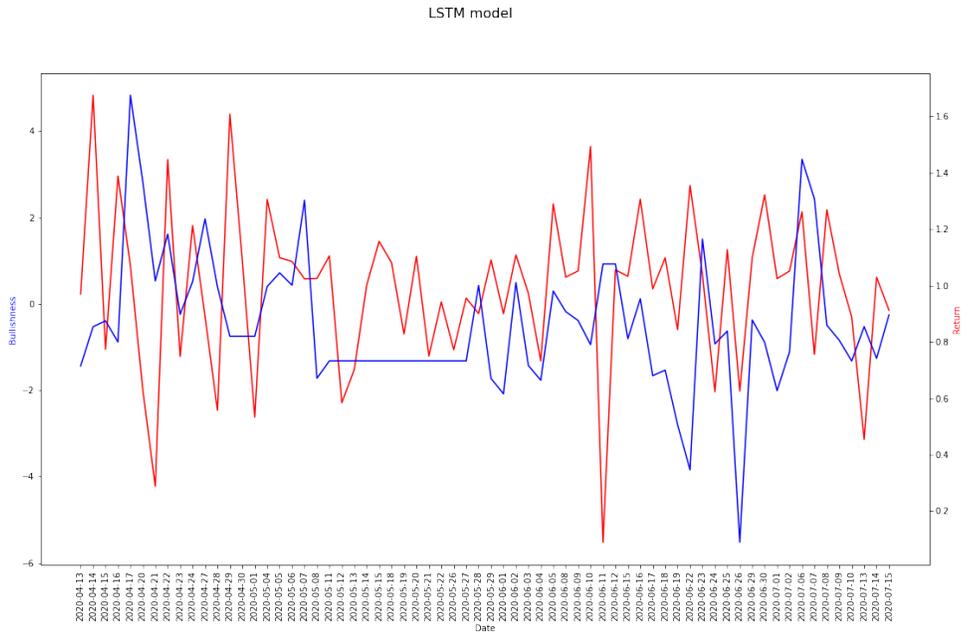

Figure 4.10 The Graph of the LSTM Model for MSFT

Figure 4.10 depicts that there are some patterns that are predicted closely in LSTM Model. However, some predictions are far away from the actual values.

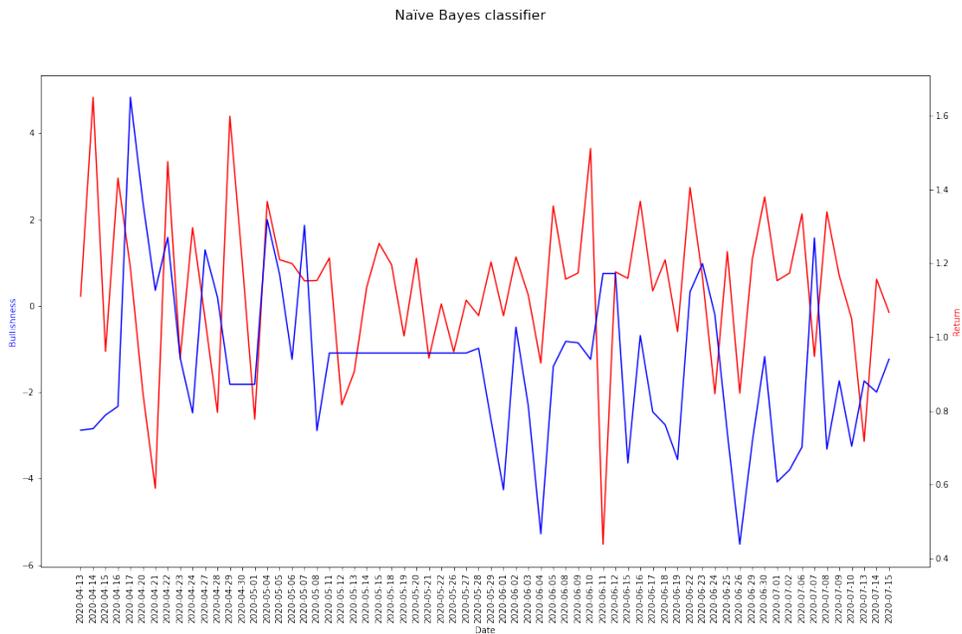

Figure 4.11 The Graph of the Naïve Bayes Classifier for MSFT

As described through Figure 4.11, most predictions of the Naïve Bayes Classifier are wrong, even concerning the positivity or negativity of the Return.

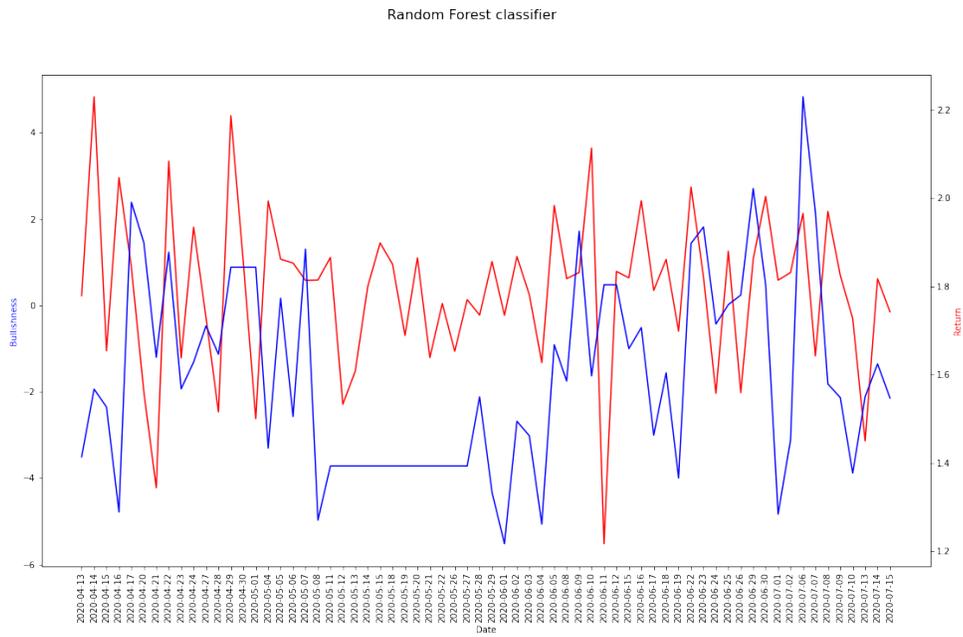

Figure 4.12 The Graph of the Random Forest Classifier for MSFT

The predictions of the Random Forest classifier are not that satisfactory Figure 4.12, but closer than the other two.

## 4.2.4   Evaluation of the Tesla Stock ($TSLA)

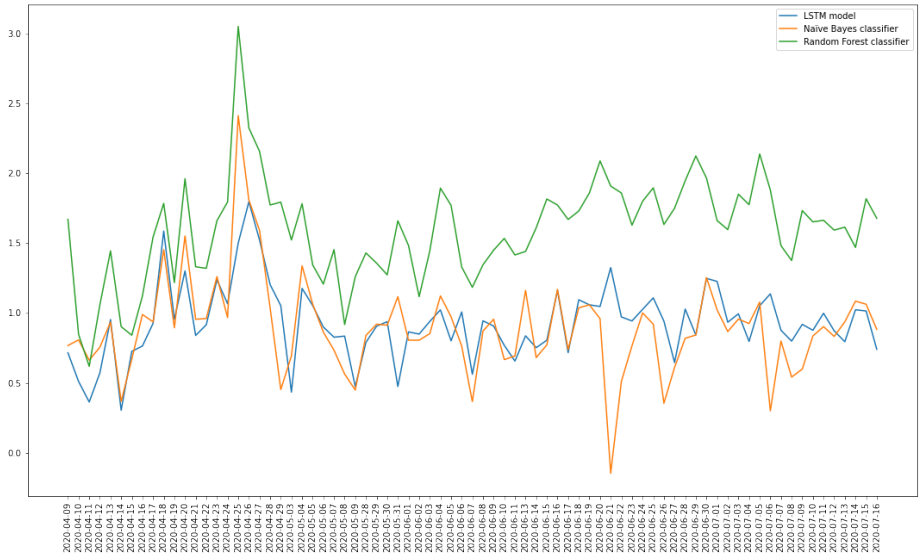

Figure 4.13 The Graph of TSLA Bullishness

Considering Figure 4.13, Naïve Bayes and LSTM outputs are more similar compared with Random Forest. Many similarities between the patterns can be observed in all three. All classifiers agree on the peak points.

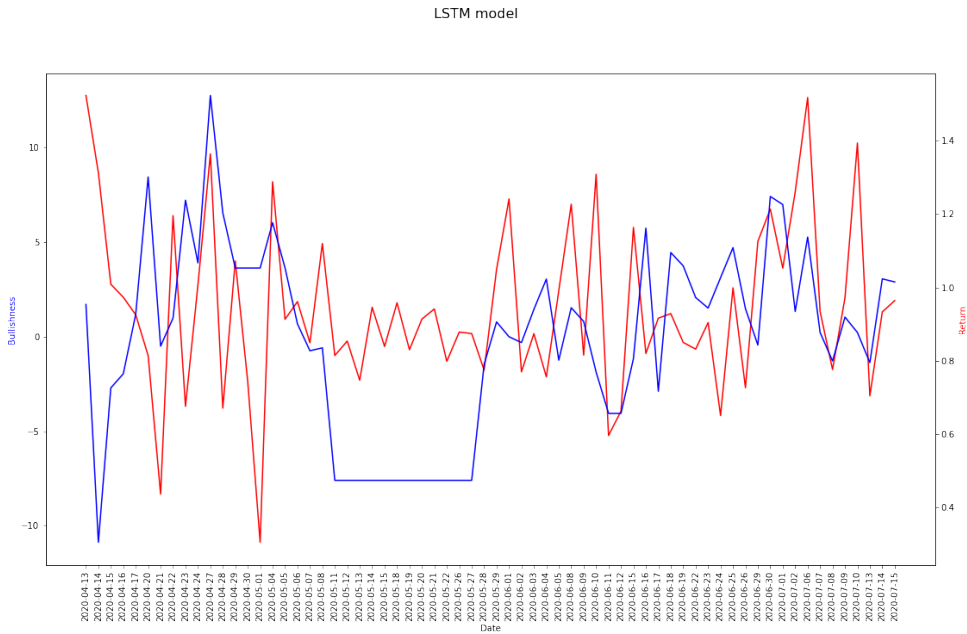

Figure 4.14 The Graph of the LSTM Model for TSLA

There are a few patterns that are predicted closely in LSTM model (Figure 4.14). However, most predictions are far from the actual values.

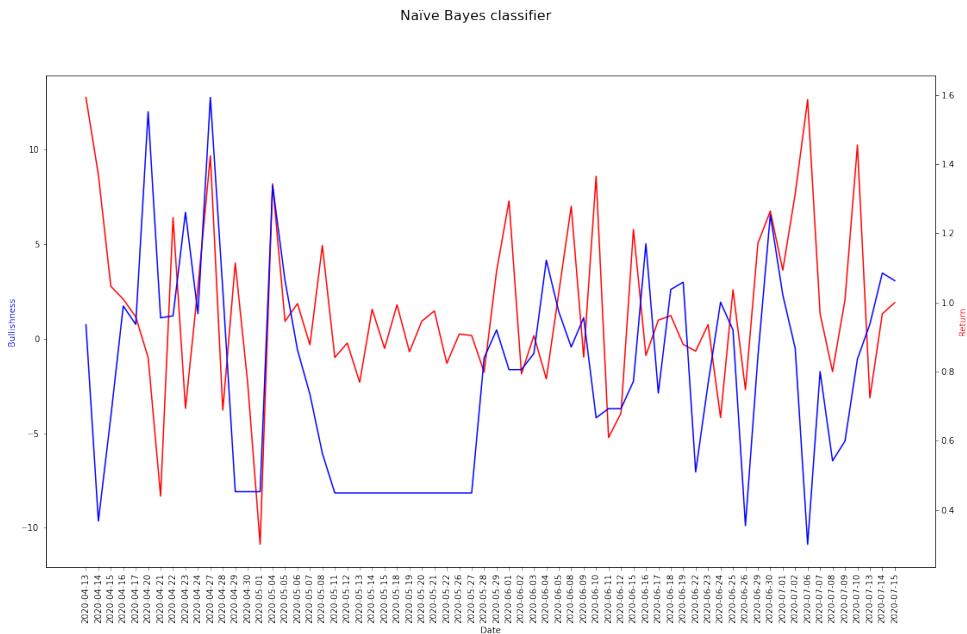

Figure 4.15 The Graph of the Naïve Bayes Classifier for TSLA

By analyzing (Figure 4.15), we can understand that the Naïve Bayes has done better than LSTM in this case, most of the peaks and minimums are predicted accurately.

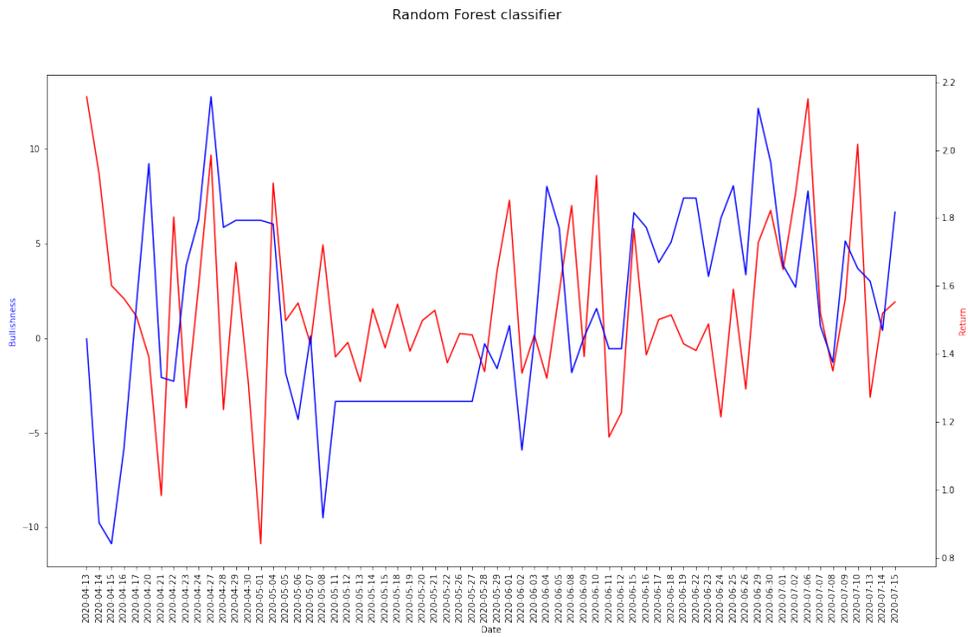

Figure 4.16 The Graph of the Random Forest Classifier for TSLA

Random Forest predictions are the closest (Figure 4.16), and this diagram has the closest predictions of all, with only a few errors.

## 4.2.5   Evaluation of the Amazon Stock ($AMZN)

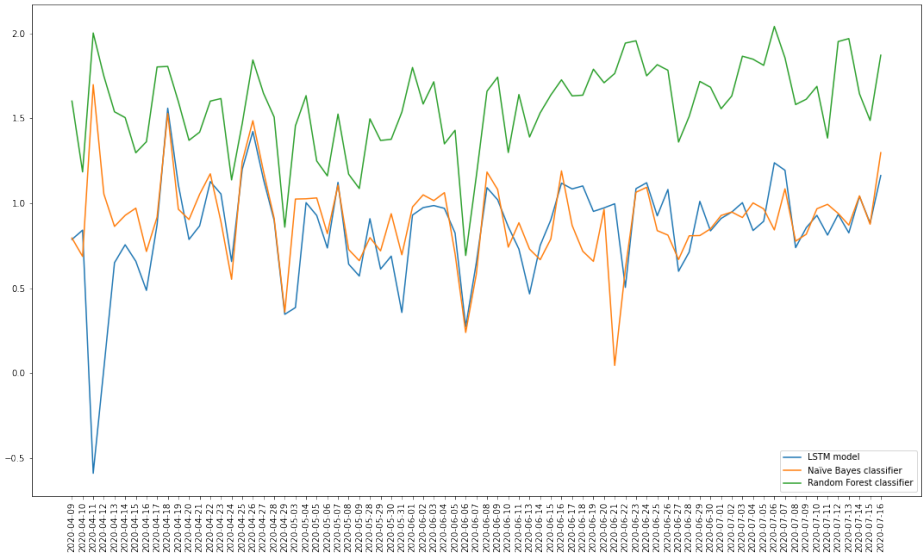

Figure 4.17 The Graph of AMZN Bullishness

The three models agree on most of the rising or falling patterns as shown in Figure 4.17. Random Forest has predicted the most optimistic and Naïve Bayes the most pessimistic. LSTM is somewhere in between.

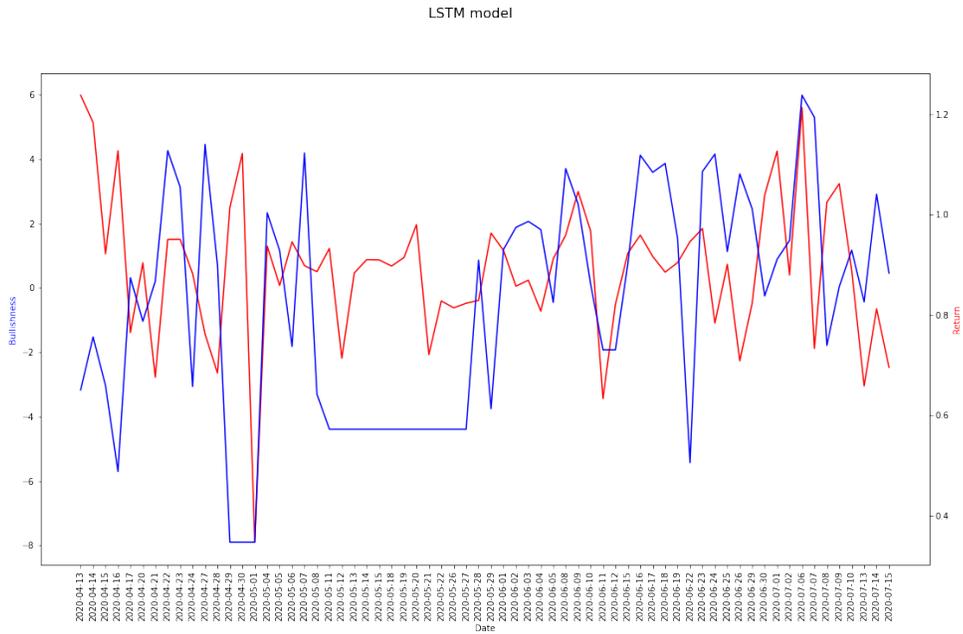

Figure 4.18 The Graph of LSTM Model for AMZN

There are some patterns that are predicted very closely in LSTM model (Figure 4.18). However, some predictions are far away from the actual values.

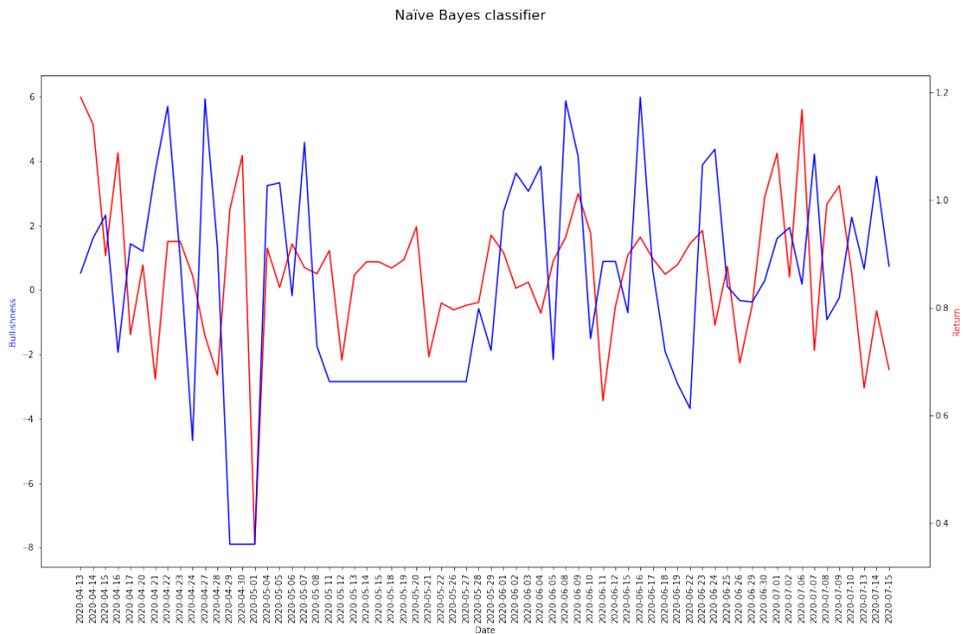

Figure 4.19 The Graph of the Naïve Bayes Classifier for AMZN

The predictions of Naïve Bayes are not that satisfactory as it seems in Figure 4.19, yet still better than LSTM's.

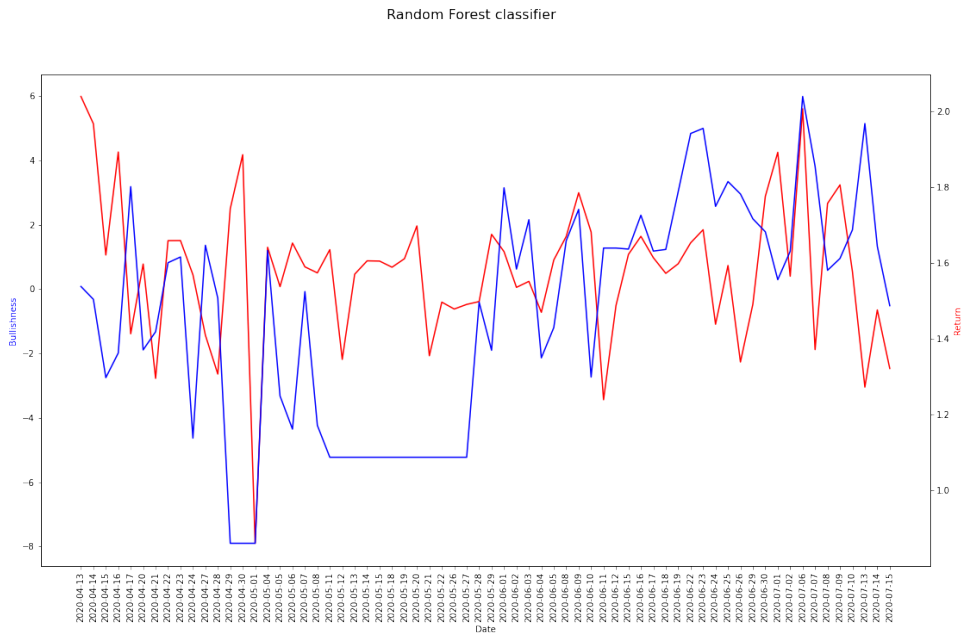

Figure 4.20 The Graph of the Random Forest Classifier for AMZN

Figure 4.20 depicts that the Random Forest's performance is similar to Naïve Bayes, on some days Naïve Bayes' results are closer, and on some Random Forest's.

## 4.2.6 Overall Comparison

Random Forest predictions are the closest of all. LSTM and Naïve Bayes performances were quite similar. Sometimes LSTM did better, and in some cases, Naïve Bayes outperformed. Considering all cases, Naïve Bayes did slightly better. We must take into account that our parameters were not set with exact care in building the models, along with other possible errors in data or parameter calculation.

# Chapter 5

# Conclusion

The key results of this dissertation, along with potential future research directions, are outlined in this chapter.

## 5.1   Results and Discussions

Many outstanding pieces of research have been reviewed in the increasingly popular field of sentiment analysis, and a variety of cutting-edge works have been studied in this document.

Here, we predict how the US stock market will move in the future by analyzing the polarity of Twitter posts about the market. We presented various stages and strategies in each sentiment analysis technique. Additionally, we looked into the several forms of data and tools that might have been used in the sentiment analysis study and offered suggestions for their advantages and disadvantages. In general, this study seeks to determine whether there is any connection between public opinion ratings and stock price changes.

This study also discusses identifying the most appropriate sentiment methodologies to utilize in such a scenario and determining the most relevant variables for predicting stock price fluctuations. However, filtering and processing the dataset in the data portion proved difficult and time-consuming. For instance, a significant portion of tweets occasionally contained spam. Therefore, we also spent considerable time labeling and tokenizing tweets.

After the data had been correctly organized and cleaned, various sentiment analysis models and algorithms were used, and their evaluation was conducted.

Specifically, we examined the average sentiment values of this Twitter dataset using Long Short-Term Memory, Bernoulli Naïve Bayes, and Random Forest classifiers. Then, we compared the results to see which model was the most accurate, showing that all these algorithms correlate.

Finally, the primary assessment initiatives were summarized in chapter three, along with a thorough analysis of our model's outcomes for the five stock symbols.

According to the outcome of stocks' evaluation in chapter 4, the sentiments of tweets significantly impact share prices, and the results of the algorithms have demonstrated a meaningful influence on the trends in the financial markets.

In other words, we concluded that tweets about a specific company might affect its share price performance. Of course, by stating this, we are not claiming an always direct connection, but data has shown that tweet sentiments often cause fluctuations in stock prices and market cap.

## 5.2   Prospects and Future work

Although the overall goals of the dissertation were met, there are still a few areas that may be improved. It would be intriguing to see this research expanded upon in order to enhance a sentiment indicator used as a momentum signal for the stock market.

In other words, we think there is room for improvement in the precision of these machine learning methods. Here is a list of suggestions that we believe could be helpful in this regard.

- Although Twitter has many international users, we solely concentrate on English sentences. Our method should work to categorize sentiment in other languages as well.

- Neutral tweets cannot be disregarded; neutral sentiment requires adequate consideration. To obtain more accurate results, the same criteria must be taken into account for audio files, movies, and photos.

- Current algorithms categorize the general tone of a tweet. The polarity of a tweet can vary depending on your interpretation of the tweet. For example, the sentiment in the tweet "England beats Germany" is favorable for England and unfavorable for Belgium. Semantics could be useful here. The classification would be done in accordance with which noun is primarily connected with the verb based on the results of a semantic role labeler. This may allow "Germany beats England" to be categorized differently as "England beats Germany."

- It is important to extend the study period to capture a wider variety of tweets and events, overcoming the limitations of specific time events like the COVID-19 pandemic. The productivity of the models may also be better understood by examining them in the context of various economic conditions, such as booms and recessions.

- An artificial neural network can be used to analyze sentimental tweets more accurately.